\documentclass{ieeeconf} 
\IEEEoverridecommandlockouts 

\usepackage{amsmath}
\usepackage{amssymb}
\usepackage{caption}
\usepackage{subcaption}
\usepackage{graphicx}
\usepackage{xcolor}

\usepackage{lipsum}  
\usepackage{hyperref}
\usepackage{bm}
\usepackage[export]{adjustbox}
\title{Computational Design of Magnetic Soft Shape-Forming Catheters using the Material Point Method}

\author{
        ~Joshua~Davy, ~Peter~Lloyd, ~James~H.~Chandler \textit{Member, IEEE} and Pietro Valdastri \textit{Fellow, IEEE} % <-this % stops a space
\thanks{~Joshua~Davy,
        ~Peter~Lloyd,
        ~James~H.~Chandler,
        ~and~Pietro~Valdastri are with the STORM Lab, Institute of Autonomous Systems and Sensing (IRASS), School of Electronic and Electrical Engineering, University of Leeds, Leeds, UK.
Email: \{\tt{el17jd, men9prl,  j.h.chandler, p.valdastri}\}@leeds.ac.uk}
}

\begin{document}
 
\maketitle

\begin{abstract}
Magnetic Soft Catheters (MSCs) are capable of miniaturization due to the use of an external magnetic field for actuation. Through careful design of the magnetic elements within the MSC and the external magnetic field, the shape along the full length of the catheter can be precisely controlled. However, modeling of the magnetic-soft material is challenging due to the complex relationship between magnetic and elastic stresses within the material. Approaches based on traditional Finite Element Methods (FEM) lead to high computation time and rely on proprietary implementations. In this work, we showcase the use of our recently presented open-source simulation framework based on the Material Point Method (MPM) for the computational design of magnetic soft  catheters to realize arbitrary shapes in 3D, and to facilitate follow-the-leader shape-forming insertion. 
\end{abstract}

\IEEEpeerreviewmaketitle

\section{Introduction}
\label{sec:intro}
 Magnetic Soft Catheters (MSCs) have been proposed as a solution for minimally invasive medical procedures \cite{zhou2021ferromagnetic}. The ability to achieve miniaturization and precise shape control of these catheters using an external magnetic field has opened up new possibilities for targeted interventions \cite{pittiglio2023personalized}. Through careful design of magnetic elements integrated within the catheter, the full-length shape can be controlled to navigate through intricate anatomical pathways with enhanced accuracy and dexterity. This, along with the soft structure of the catheter, reduces the anatomical forces on the surrounding tissues with the potential to reduce patient trauma \cite{lloyd2022magnetic}.

 However, the successful implementation of MSCs hinges upon the ability to accurately model the complex interaction between magnetic and elastic stresses within the catheter's soft material. Traditional approaches, such as Finite Element Methods (FEM), have been employed to simulate and analyze these interactions \cite{lloyd2020optimal}. Yet these methods often suffer from high time complexity and rely on proprietary software/toolkits limiting accessibility and hindering broader research efforts.

 In our recent work, we presented a simulation framework for modelling of magnetic soft materials based on the Material Point Method (MPM) \cite{davy2023framework}. Unlike previously presented FEM approaches, our framework inherently models self-collision between areas of the model and can capture the effect of forces in non-homogeneous magnetic fields. We demonstrated this framework on various test cases but also on robot designs presented in the literature with good agreement with the authors' real-world results.

 Previous work on this, based on FEM analysis was found to have high time complexities limiting the ability to apply iterative optimization processes to these models \cite{Lloyd2020}. Instead a data-driven approach was applied where a data-set of MSC deformations under various applied fields was initially created and a neural network trained to approximate the inverse function. Although functional, this system was limited to a 2D case and three magnetic elements. Data-driven approaches often fail to generalize far from their provided data-set. For this approach to move to 3D catheters under a wider range of deformations, the required data collection and training process would become computationally intractable. In contrast, FEM based approaches have been directly optimized on using evolutionary algorithms \cite{wang2021evolutionary}, however the combination of large computation time and high number of iterations for convergence can lead to large run-times.

\begin{figure*}[th!]
 \centering
 \includegraphics[width=0.84\textwidth]{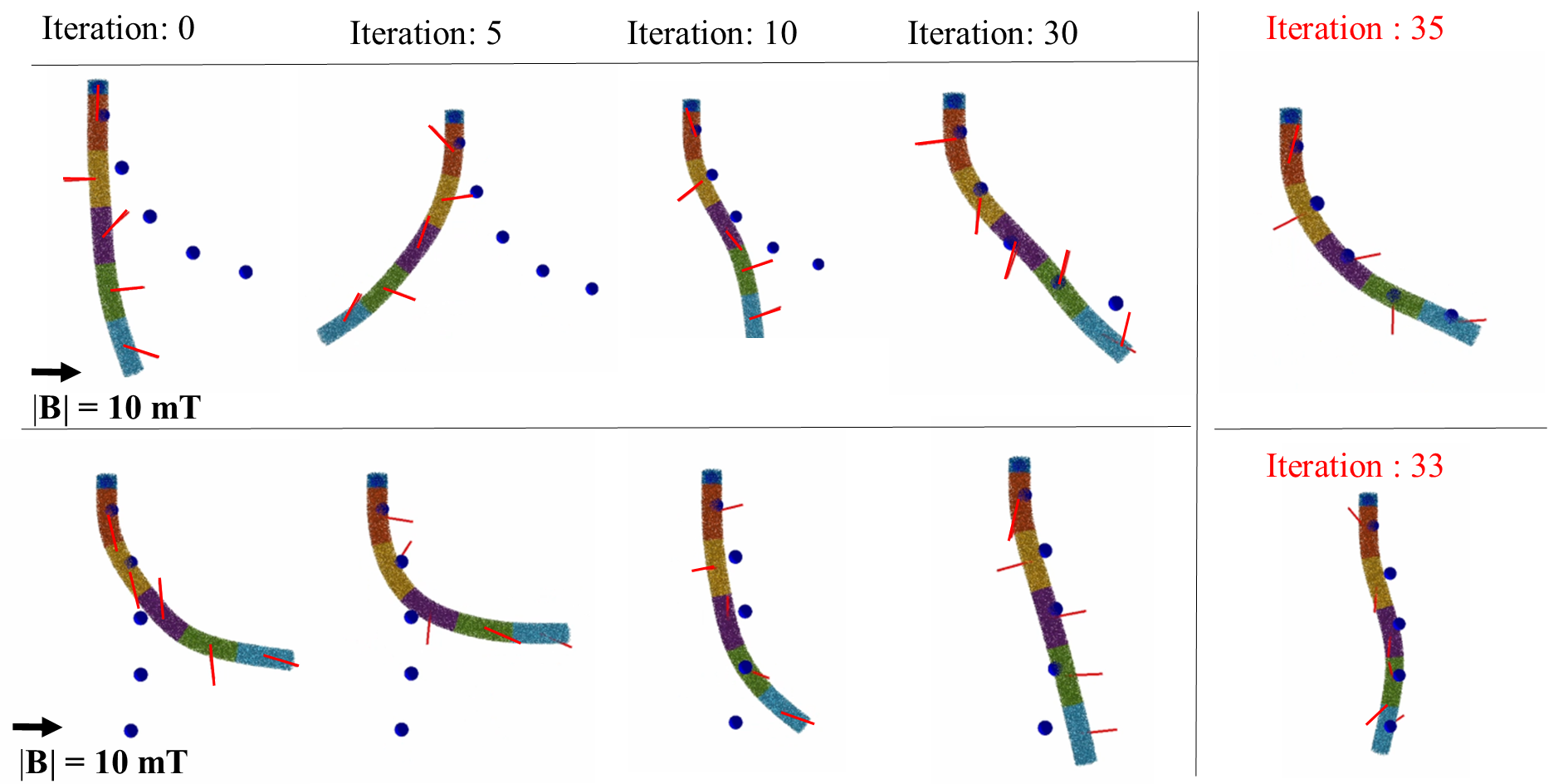}
 \caption{Convergence of static deformations of MSCs. Target shapes shown as blue dots and magnetization vectors by red arrows. The actuating field is fixed and the magnetization vectors are optimized to minimze the gap between deformation and target shape.}
 \label{fig:static}
\end{figure*}

Another approach has been to utilize more simplified modelling such as rigid links \cite{lloyd2022magnetic}. However, this fails to capture the full geometric structure of the MSC and introduce additional dependent variables such as number of links, link length etc.. Further, the twisting instability of MSCs around their major-axis has been noted as an important factor in their design \cite{lloyd2022magnetic}. Simplified models fail to capture this. 

In this work,  we consider the application of the \textit{magneticMPM} framework for computational design of shape-forming catheters. In particular, we showcase the ability to model the 3D deformation of catheters in dynamic simulation. Our computational design approach is computationally tractable with each iteration lasting between 20 and 25 seconds. The use of a surrogate based optimization approach allows convergence of our process within 100 iterations.

\section{Simulation of catheters with magneticmpm}
\label{sec:sim}

Previously, we presented an open-source simulation framework for representing magnetic soft robots based on MPM. The magneticMPM framework is available at \url{https://github.com/joshDavy1/magneticMPM}. The simulation framework is capable of representing magnetic soft robot designs of various scales and geometry. In particular, we replicated the designs of \cite{xu2019} and \cite{hu2018nature} in simulation. MPM utilizes a dual particle-grid representation of the material. The lack of a deforming underlying mesh (Such as in FEM), allows large deformations to be represented without the risk of mesh collapse. 
The MSCs we consider are fully soft and are formed of silicone polymers mixed with NdFeB magnetic powder. Once cast and cured, these catheters are then subjected to an impulse magnetic field that sets the direction of remnant magnetization. The direction of this magnetization as a function of the robot's length is referred to as it's magnetic profile. This magnetic profile can be carefully designed in order to give desired deformations under an applied magnetic field.

To simulate the MSC in magneticMPM, we adopt the Neo-Hookean hyper-elastic model to represent the soft material in Piola-Kirchoff form

\begin{equation}
    \textbf{P}^t_{p,elastic} = GJ^{-2/3}(\textbf{F}_p^t- \frac{I_1}{3}\textbf{F}^{-T}) + KJ(J-1)\textbf{F}^{-T}.
    \label{eqn:neoHooke}
\end{equation}

where $\textbf{P}p^t_{p,elastic}$ is the elastic contribution of the first Piola-Kirchoff stress tensor, $G$ is the shear modulus of the material and $K$ is the bulk modulus. $J$ = $\text{determinant}(\textbf{F}_p^t)$ and $I_1$ = $\text{trace}(\textbf{F}_p^{t^T}\textbf{F}_p^t)$. Assuming near incompressiblity of the utilized silicone polymers we chose the bulk modulus to be a sufficiently high value. In this case we set $K$ = $20G$.

For the magnetic properties, we use the stress-relationship derived in Zhao et. al. \cite{Zhao2019}

\begin{equation}
    \textbf{P}^t_{p,magnetic} = - \frac{1}{\mu_0}\textbf{B} \otimes \textbf{B}_{r_p}, 
\end{equation}
where $\otimes$ is the dyadic product and $\textbf{B}_{r_p}$ is the remnant magnetic flux density associated with the particle. $B$ is the applied field.

\section{Computational Design}
\label{sec:design}
By iterative optimization, the magnetic profile of the simulated MCR can be adjusted until the desired deformation is achieved. To achieve this, we can formulate the optimization as a minimization process using the cost function

\begin{equation}
    f(\boldsymbol{\theta}) =  \sum_i^N  ||\mathbf{p}_{i} -\mathbf{\hat{p}}_{i}||^2.
\end{equation}

where $\boldsymbol{\theta}$ is the magnetic profile, $\mathbf{p}_{i}$ is the desired position at distance $i$ along the robot $i \in \{0, N\}$. $\mathbf{\hat{p}}_{i}$ is the equivalent resultant position in simulation. This cost is then minimized in order to give the final magnetic profile.

The optimization process can be treated as an expensive black-box optimization problem. With the computational time of each iteration being in the range of 20 - 25 seconds. We apply the Bayesian optimization algorithm to this problem in order to converge in a tractable time. This, algorithm relies on a underlying surrogate function that is sampled in order to choose the next magnetic profile. Each iteration is utilized to improve this surrogate function. More details can be found in \cite{frazier2018tutorial}.

\begin{figure*}[th!]
 \centering
 \includegraphics[width=0.85\textwidth]{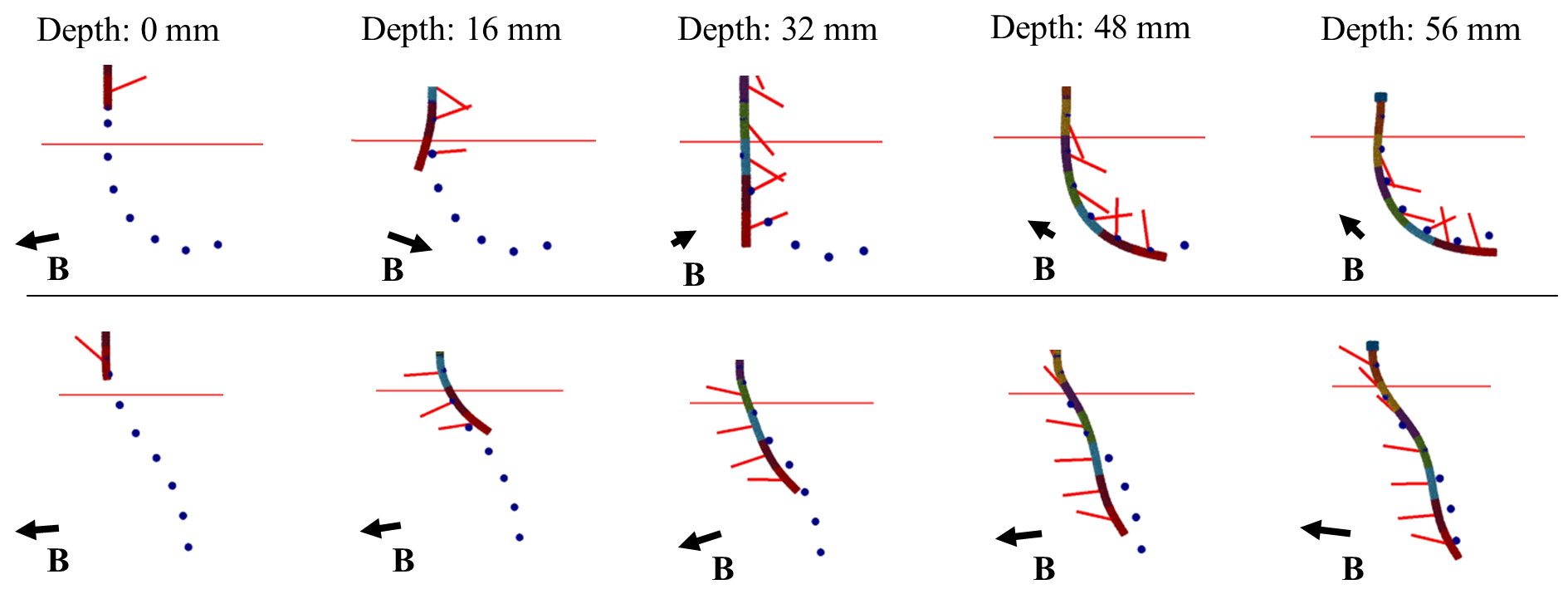}
 \caption{Converged solutions of two MSCs under follow-the-leader actuation.}
 \label{fig:insertion}
\end{figure*}

\section{Results}
\label{sec:experiments}
To evaluate our methodology, we assume the material properties of previously presented MSCs shown in \cite{pittiglio2022}. Each MSC is formed of $N$ magnetic segments each with their own direction of magnetization. An external homogenous magnetic field is applied leading the MSC to deform.

\subsection{Static deformation}
We first consider the static deformations of the MSC under a fixed applied field with $N = 5$. The catheter length is $40$ mm with diameter $4$ mm. The optimization process was repeated until an average RMS error of below $2$ mm between target and desired shape was achieved. Figure \ref{fig:static} shows two examples of the iterative process on two arbitrary target shapes. Convergence was achieved in $35$ and $33$ iterations respectively. From random initialization, it can be observed that the solution is found in a small number of iterations.

\subsection{Deformation under continuous insertion}
When considering a MSC inserting into a lumen, a follow-the-leader trajectory must be followed. Here, a target path is defined which the catheter must follow whilst inserted into the space. The catheter length is $56$ mm and diameter $3$ mm. With the need to change the sectional deformation as the catheter is inserted, the applied field is also included in the optimization vector. The field norm was limited to a maximum $B = 10$ mT. Figure \ref{fig:insertion} shows two examples of converged solutions for arbitrary shapes. Both converged in under a $100$ iterations with an exit error of $3$ mm.

\section{Conclusion}
\label{sec:conclusion}
In this work, we consider the computational design of magnetic soft shape-forming catheters. Utilizing our open-source framework, we have successfully shown the tractability of 3D simulations of MSCs and integrated that into an optimization problem in order to solve for shape-forming tasks. Similar work \cite{lloyd2020optimal}, based on evolutionary algorithms require hundreds of iterations before convergence. The underlying surrogate function in Bayesian optimization allows efficient sampling with the simulation keeping the runtime within tractability.

Compared to the data driven approach in \cite{lloyd2020optimal}, our proposed model-based optimization generalizes well to various deformations and robot geometries. This avoids any need for large dataset collection and training process. However, to be an accurate representation of real world MSCs, the underlying simulation material parameters must closely match. In this work, we relied on existing material parameter data from previous work \cite{veiga2021}, however these optimal parameters will depend on the strain ranges experienced.

The use of a surrogate based optimization with a realistic simulation model could be considered a hybrid between data-driven and model-based approaches. The surrogate model is incrementally improved with each sample of the real world simulation and is used to suggest new sample points for the relatively expensive simulation model. 

Although our previous work showcased the strong agreement between our magneticMPM framework and real world results \cite{davy2023framework}, the results shown in this work must be validated against real world fabricated MSCs. We will consider how this methodology can be further applied to optimize additional factors, such as catheter geometry, energy minimization and to achieve dynamic time-variant behaviours.

% \ifCLASSOPTIONcaptionsoff
%   \newpage
% \fi

\bibliographystyle{IEEEtran} 
\bibliography{references}{}

\end{document}